\documentclass[lettersize,journal]{IEEEtran}
\usepackage{amsmath,amsfonts}
\usepackage[ruled,vlined]{algorithm2e}
\usepackage{algpseudocode}
\usepackage{array}
\usepackage[caption=false,font=normalsize,labelfont=sf,textfont=sf]{subfig}
\usepackage{textcomp}
\usepackage{stfloats}
\usepackage{url}
\usepackage{verbatim}
\usepackage{graphicx}
\usepackage{cite}
\usepackage{amsmath}
\usepackage{amssymb}
\usepackage{bm}
\usepackage{multirow}

\usepackage[capitalize]{cleveref}
\crefname{section}{Sec.}{Secs.}
\Crefname{section}{Section}{Sections}
\Crefname{table}{Table}{Tables}
\crefname{table}{Tab.}{Tabs.}

\usepackage{xspace}
\makeatletter
\DeclareRobustCommand\onedot{\futurelet\@let@token\@onedot}
\def\@onedot{\ifx\@let@token.\else.\null\fi\xspace}
\def\etal{\emph{et al}\onedot}
\def\etc{\emph{etc}\onedot}
\makeatother

\begin{document}

\title{Spatio-Temporal Relation Learning for Video Anomaly Detection}

\author{Hui Lv, Zhen Cui\thanks{Corresponding author: Zhen Cui, zhen.cui@njust.edu.cn.}, Biao Wang, Jian Yang
\thanks{Email address: (hubrthui, csjyang)@njust.edu.cn (H. Lv, J. Yang).}
\thanks{H. Lv, Z. Cui and J. Yang are from School of Computer Science and Engineering, Nanjing University of Science and Technology, Nanjing, Jiangsu, China.}
}


\markboth{Journal of \LaTeX\ Class Files,~Vol.~14, No.~8, August~2021}%
{Shell \MakeLowercase{\textit{et al.}}: A Sample Article Using IEEEtran.cls for IEEE Journals}


\maketitle

\begin{abstract}
Anomaly identification is highly dependent on the relationship between the object and the scene, as different/same object actions in same/different scenes may lead to various degrees of normality and anomaly. Therefore, object-scene relation actually plays a crucial role in anomaly detection but is inadequately explored in previous works. In this paper, we propose a Spatial-Temporal Relation Learning (STRL) framework to tackle the video anomaly detection task. First, considering dynamic characteristics of the objects as well as scene areas, we construct a Spatio-Temporal Auto-Encoder (STAE) to jointly exploit spatial and temporal evolution patterns for representation learning. For better pattern extraction, two decoding branches are designed in the STAE module, i.e. an appearance branch capturing spatial cues by directly predicting the next frame, and a motion branch focusing on modeling the dynamics via optical flow prediction. Then, to well concretize the object-scene relation, a Relation Learning (RL) module is devised to analyze and summarize the normal relations by introducing the Knowledge Graph Embedding methodology. Specifically in this process, the plausibility of object-scene relation is measured by jointly modeling object/scene features and optimizable object-scene relation maps. Extensive experiments are conducted on three public datasets, and the superior performance over the state-of-the-art methods demonstrates the effectiveness of our method. 
\end{abstract}


\section{Introduction}
\label{sec:intro}
\IEEEPARstart{V}{ideo} anomaly detection (VAD) refers to the identification of events that conform to unexpected behavior/activity, and is with increasing demand in security surveillance and monitoring. In many cases, abnormal events rarely occur in real-world videos, and manual checking from massive video data would be significantly tedious and time-consuming. Therefore, automatic abnormal events detection is pretty desirable for liberating humans from exhausting efforts of video surveillance. Consequently, a growing interest~\cite{ramachandra2020survey,kiran2018overview,chandola2009anomaly,santhosh2020anomaly} has been recently dedicated to the research topic of VAD.

\begin{figure}[t]
	\centering
	\includegraphics[width= 0.45\textwidth]{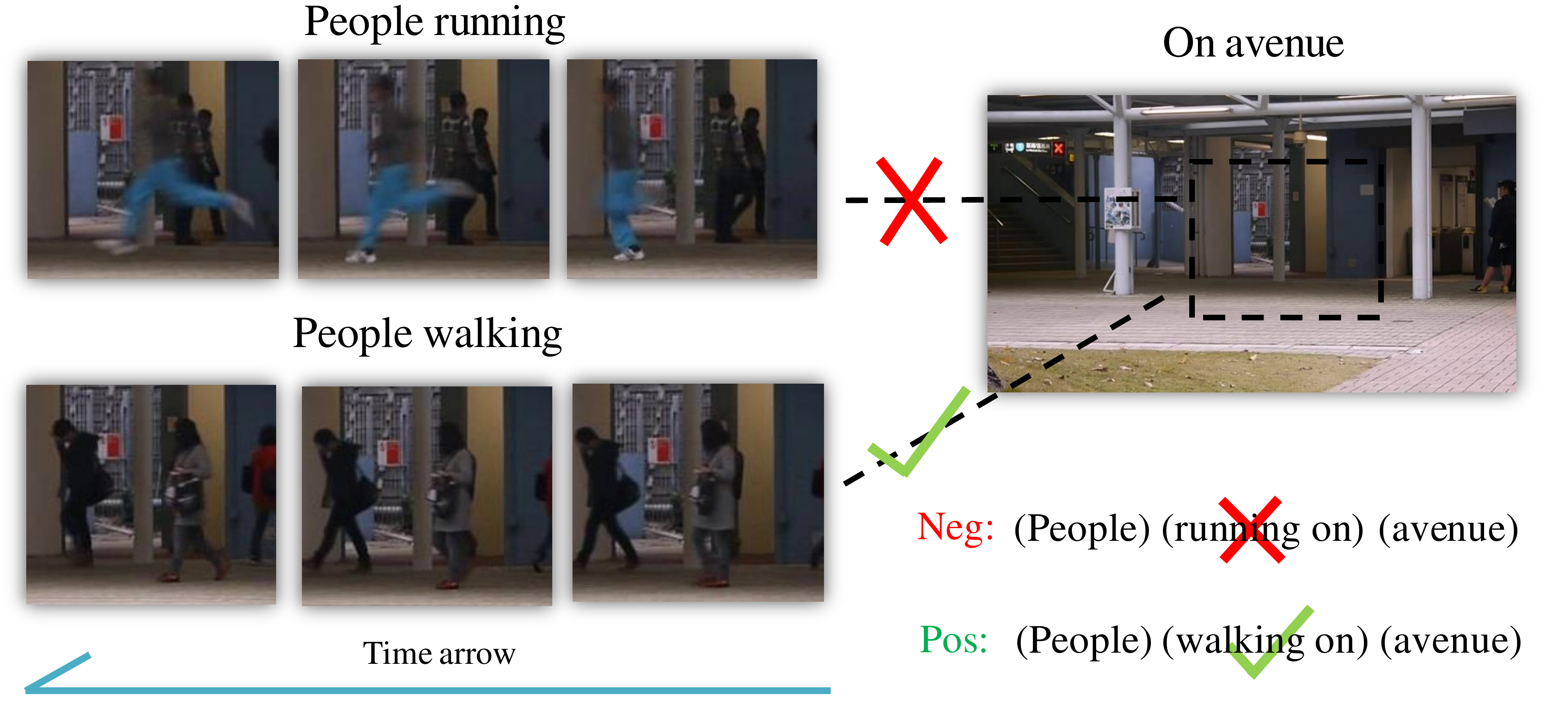}
	\caption{An illustration of object-scene relation modeling in anomaly detection. 
	On avenue, the fact of `People walking on avenue' is positive (normal) while `People running on avenue' is negative (anomalous). Anomalous status of the object-scene tuple: (people-avenue) is determined by the relation therein. The observed relation (walking on) is normal and any unseen relation (running on) tends to be anomaly. }
	\label{fig:abstruct}
\end{figure}

VAD is an open and challenging problem as the `anomaly' is extremely rare, conceptually unbounded, and often ambiguous, making it infeasible to gather a large quantity and all kinds of possible anomalous data. 
A typical solution is to formulate the problem as an unsupervised one, by fitting a model to exploit the regular patterns on normal data. Then the events that deviate from the model are considered as anomalies. Benefiting from the spurt of deep learning algorithms, nowadays great success has been witnessed~\cite{hasan2016learning,luo2017revisit,liu2018future,nguyen2019anomaly,gong2019memorizing,park2020learning,lv2021localizing,lv2021learning} in computer vision tasks. A similar trend also exists in the research academic of anomaly detection. 
A lot of deep neural network based methods have been proposed to tackle the problems therein, which exceed the traditional handcrafted feature based methods in the aspect of either speed or accuracy.
To date, reconstruction or future prediction is a popular paradigm in the era of deep VAD methods. 
Researchers usually adopt a Deep Auto-Encoder (DAE) to learn the normal patterns based on historical observation and to reconstruct the current frame \cite{hasan2016learning,masci2011stacked,sabokrou2016video,chalapathy2017robust,sabokrou2018adversarially,abati2019latent} or predict the upcoming frame \cite{liu2018future,nguyen2019anomaly,lu2019future,luo2017remembering,gong2019memorizing,liu2021hybrid}.
Typically, they assume that the normal pattern in the context is automatically encoded and well exploited during the training phase of DAE.
Therefore, in the testing phase, if a frame agrees with its prediction, it potentially corresponds to a normal event. Otherwise, it potentially corresponds to an anomaly.

Considerable progress of VAD has been witnessed through reconstruction or future prediction frameworks, especially on datasets with simple scenes like Ped2~\cite{liu2018future,nguyen2019anomaly,lv2021learning}.
However, two critical issues still need to be tackled to further boost the VAD task. One fundamental problem is about the relation learning between objects and scenes because anomaly identification is highly scene-dependent. For instance, the same running action of one person may be always normal in the playground, but anomalous on the walking avenue. A clear illustration is shown in Figure~\ref{fig:abstruct}. However, previous works focus more on the evolution patterns of the objects themselves while failing to adequately model the crucial object-scene relation/interaction. The other problem is to strengthen spatio-temporal representation ability to better depict the dynamic characteristics of both objects and scenes. For this problem, the popular auto-encoder framework pays much attention to the appearance reconstruction, while doesn't explicitly decode the motion patterns which would be helpful to promote the spatio-temporal representation.         

To tackle all the issues aforementioned, in this paper, we propose a Spatial-Temporal Relation Learning (STRL) framework to deal with the VAD task. For a given video, a Spatio-Temporal Auto-Encoder (STAE) module is first constructed to model and encode the spatial-temporal evolution patterns within a light and simple architecture. For better spatial-temporal representation ability, we design two decoding branches in the STAE module including an appearance branch directly predicting the next frame, and a motion branch explicitly modeling the dynamic evolution by predicting future optical flows~\cite{ilg2017flownet}. Then, based on the spatio-temporal representation, a Relation Learning (RL) module is specifically designed to well model the object-scene relation by introducing the Knowledge Graph Embedding methodology~\cite{hur2021survey,dettmers2018convolutional,wang2017knowledge}. In this process, the plausibility of object-scene relation, which indicates the normality between objects/events and scenes, is measured by jointly modeling object/scene features and an optimizable object-scene relation map. Finally, anomaly detection can be achieved by referring to the calculated scores of the object-scene relation plausibility. To evaluate our proposed STRL framework, extensive experiments are conducted on three public datasets and the superior performance over the state-of-the-art methods demonstrates its effectiveness.

The contributions of our STRL framework can be summarized as follows:

i) We propose a novel Spatial-Temporal Relation Learning framework which adequately considers object-scene relation to deal with the VAD task. To the best of our knowledge, this is the first work that performs object-scene relation learning by introducing the Knowledge Graph Embedding methodology. 

ii) We construct a light and simple Spatio-Temporal Auto-Encoder (STAE) module to learn better spatial-temporal representation by explicitly decoding the motion pattern as well as predicting the appearance.

iii) Our STRL model achieves the state-of-the-art (SOTA) performance on various unsupervised VAD benchmarks, which verifies the effectiveness of our model.

\section{Related Work}
\noindent \textbf{Video Anomaly Detection.}
Due to the absence of anomaly data and expensive costs of annotations, most of the recent works treat abnormal event detection as an outlier detection task, formulating the task into an unsupervised learning problem.
Typically, the researchers fit a model using only normal data during training. Then, at inference time, the events that diverge from the normality model are labeled as abnormal. 
Classic VAD tend to rely on hand crafted appearance and motion features and then build a classification model to detect anomalies~\cite{adam2008robust,antic2011video,lu2013abnormal,mahadevan2010anomaly,mehran2009abnormal}.
For instance, authors in~\cite{piciarelli2008trajectory,zhang2009learning} adopted trajectory to extract high-level features. Also, there are methods based on low-level features,such as dynamic textures~\cite{mahadevan2010anomaly}, optical flow histograms~\cite{cong2011sparse}, SIFT~\cite{cheng2015video}. To make the best of extracted features, the classic anomaly classification algorithms can be categorized into sparse coding~\cite{cong2011sparse,zhao2011online,lu2013abnormal}, markov random field~\cite{kim2009observe}, probabilistic PCA models~\cite{5206569}, \etc. 

In order to break the bottleneck of hand-crafted features in anomaly detection, there has been recent work that explores the use of deep learning approaches.
Recently, many methods leverage deep Auto-Encoder (AE) to model regular patterns and reconstruct video frames~\cite{hasan2016learning,masci2011stacked,sabokrou2016video,chalapathy2017robust,sabokrou2018adversarially,abati2019latent}. 
Hasan \etal~\cite{hasan2016learning} leveraged the reconstruction error as an estimator for abnormality.
Latter, Liu \etal~\cite{liu2018future} proposed to predict the future frame with AE and Generative Adversarial Network (GAN).
Their hypothesis is that an abnormal frame should be harder to predict than a normal one. Thus, the peak signal-to-noise ratio between the predicted frame and
the original frame is expected to be lower for abnormal frames.
To reinforce above assumption, Gong \etal (MemAE)~\cite{gong2019memorizing} and Park \etal (LMN)~\cite{park2020learning} introduced a memory mechanism for recording normal patterns among training data. 
Lv \etal (DPN)~\cite{lv2020localizing} proposed an online memory pool to dynamically learn the normal patterns along with the prediction of future frames.

Although the future prediction framework has achieved superior performances over previous reconstruction methods, there exists room for further improvement.
Above methods solely focus on the evolution patterns in time domain while neglecting the inherent relation between the objects and the scenes.
In this work, we not only model the spatio-temporal evolution patterns with the normal training data, but also exploit the relations between the objects and scenes, to accomplish a more complete detection algorithm.

\noindent \textbf{Knowledge Graph Embedding.}
A Knowledge Graph (KG) is a data structure that represents the observed facts (knowledge) in the form of relationships (edges) between entities (nodes).
Knowledge Graph Embedding (KGE) is deemed to embed components of a KG including entities and relations into continuous vector spaces, so as to simplify the manipulation while preserving the inherent structure of the KG
\cite{wang2017knowledge}. 
Generally, in a KGE approach, entities and relations are first initialized in a continuous vector space, the embeddings of them are obtained by optimizing with a designed scoring function that maximizes the plausibility of observed facts.
KGE plays a vital role in the development of applications~\cite{berant2013semantic,heck2013leveraging,damljanovic2012named,zheng2012entity,hoffmann2011knowledge,daiber2013improving,bordes2014open,bordes2014question} like semantic search, information extraction, recommendations, language understanding, question answering, as well as advanced analytics~\cite{hur2021survey}.
To the best of our knowledge, this is the first effort to introduce the KGE theory into video anomaly detection. 
In this work, we analyze and summarize the normal object-scene relations during the network training phase. 
In addition, the plausibility of the relations is measured by jointly modeling object/scene features and an optimizable object-scene relation map.

\section{Method}
\subsection{Overview}
\begin{figure*}[!t]
	\centering
	\includegraphics[width=1\textwidth,height=0.4\textheight]{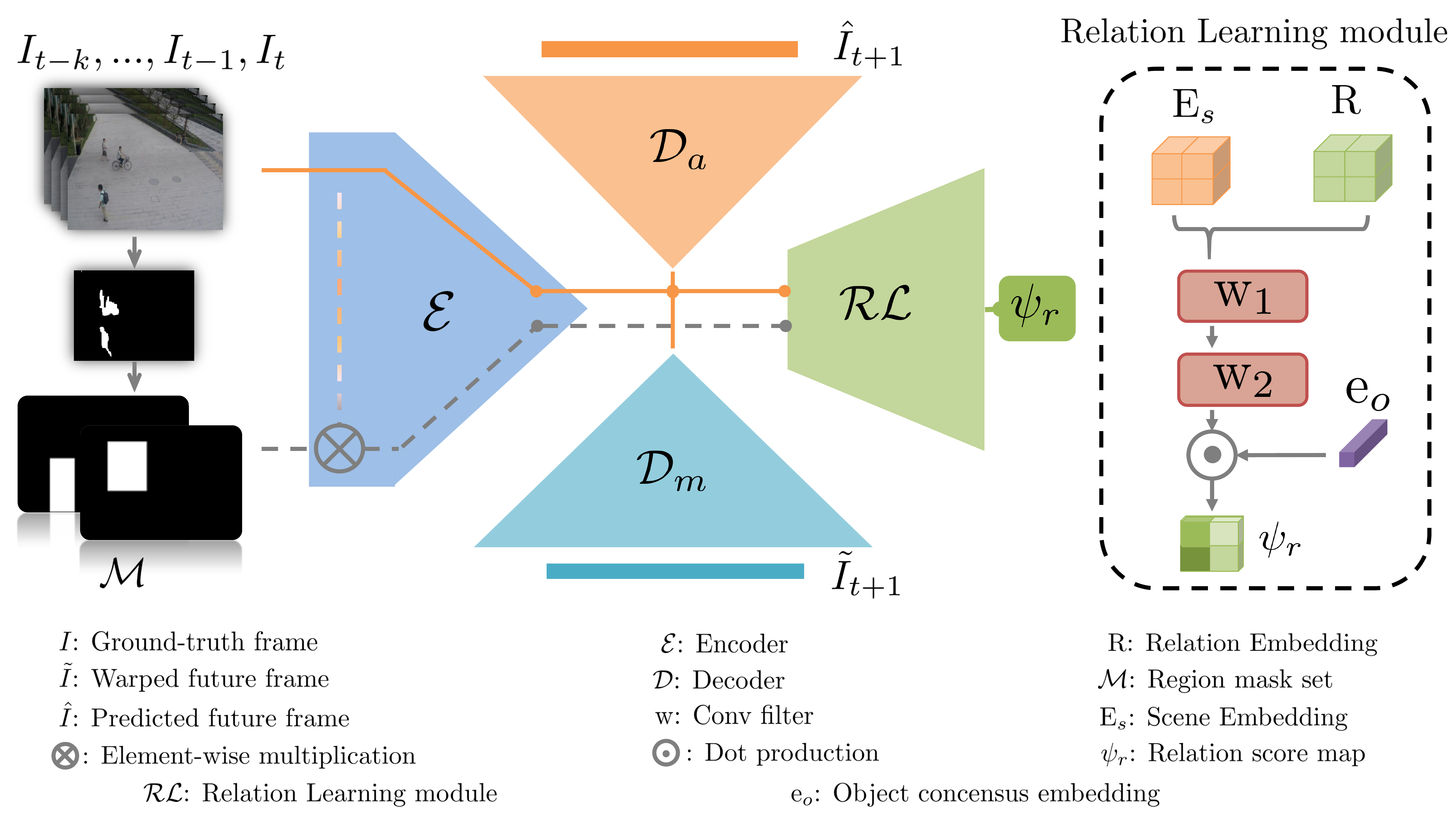}
	\caption{Overview of the STRL framework. We adopt an AE model with an encoder $\mathcal{E}$ and two encoders $\mathcal{D}_a,\mathcal{D}_m$ for learning the spatio-temporal representations to predict the future frames and optical flows. In addition, we develop a RL module $\mathcal{RL}$ to concretize and summarize the object-scene relations in the normal context. The solid line denotes the feed-forward process of data stream and the dashed line is the pass of object embedding extraction, which is not involved in the training objectives of the AE model. In the right part, we depict the forward pass in RL module. Best viewed in color.}
	\label{fig:arch}
\end{figure*}

As shown in Figure~\ref{fig:arch}, the proposed Spatio-Temporal Relation Learning (STRL) framework consists of two main components: a Spatio-Temporal Auto-Encoder (STAE) module for representation learning, together with a Relation Learning (RL) module for normal relation concretizing. The overall framework is designed to be lightweight and end-to-end trainable. 
Given a tuple of frames in a video clip, the STAE module is first constructed to encode spatio-temporal representations for predicting the RGB frames and optical flows.
With the moving area segment rule, the STAE representations are processed into object and scene embeddings.
Then, the RL module is employed to analyze and summarize the normal object-scene relations by applying the KGE method. Based on this process, the plausibility of object-scene relation is measured by jointly modeling object/scene features and an optimizable object-scene relation map.
Specifically, the plausibility scores indicate the normality between objects/events and scenes. And by combining the anomaly scores derived from the STAE module and the RL module, we can obtain a comprehensive measurement of the anomaly status.  
In this section, we first describe the learning process of normal evolution patterns in the designed STAE module in Sec.~\ref{sec_stae},
and then we present the details in the RL module in Sec.~\ref{sec_rlm}.
Finally in Sec.~\ref{sec_ad}, we detail the generation procedure of anomaly scores for temporal anomaly detecting in videos.

\subsection{Spatio-Temporal Auto-Encoder}
\label{sec_stae}

\begin{figure}[t]
	\centering
	\includegraphics[width=0.48\textwidth]{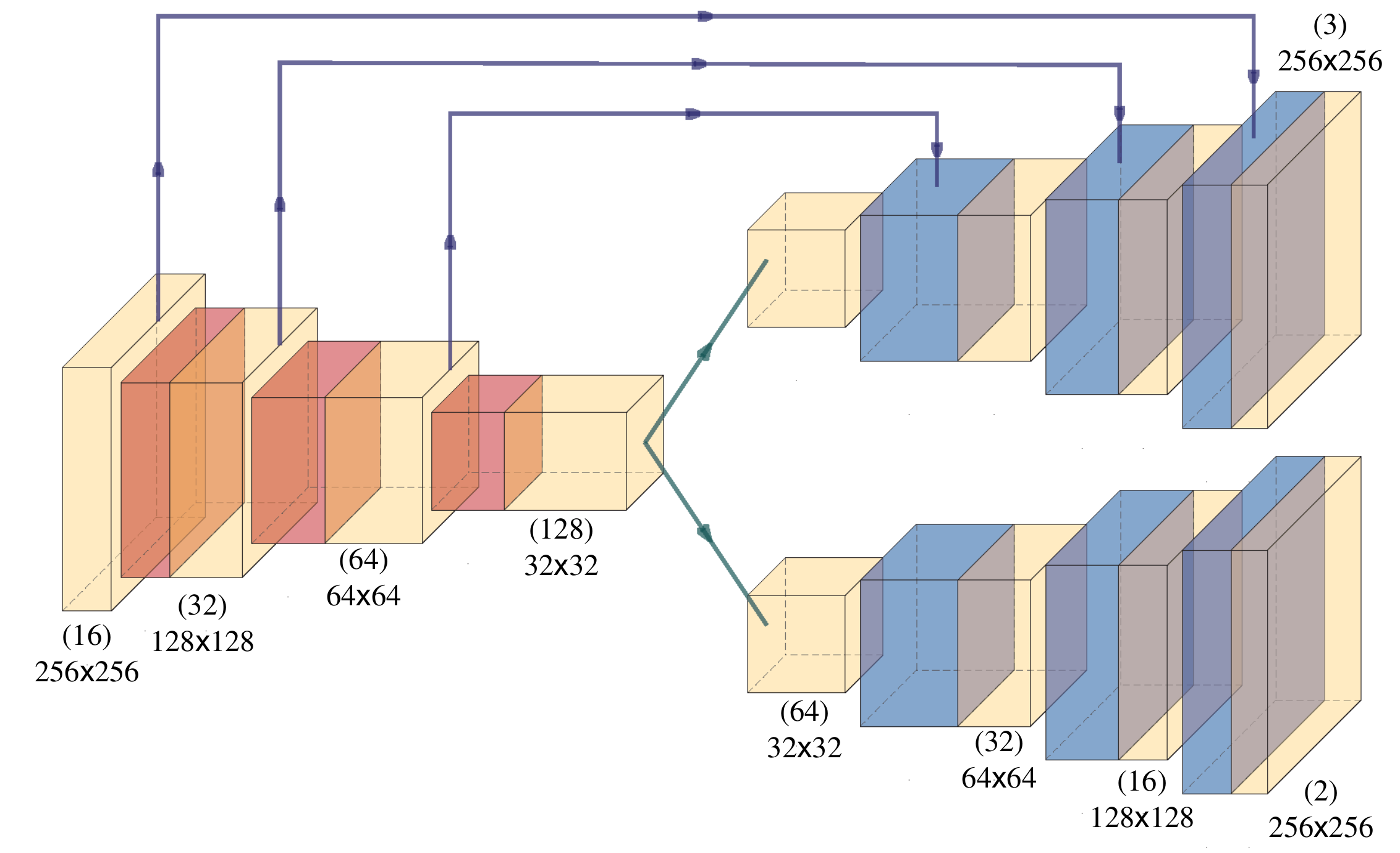}
	\caption{Overview of The STAE module. Our model contains one encoder and two decoders, with the top-right one with skip connections for decoding the upcoming frames ($3$-channel) and the bottom-right one for predicting the future optical flows ($2$-channel). We downsample the spatial resolution for $3$ times. The light-orange, dark-orange, blue blocks denote the Conv, Strided-Conv and Transpose-Conv layers, separately. Each layer is followed by a BN and ReLU operation. Best viewed in color.}
	\label{fig:STAE}
\end{figure}

The designed video anomalous detection (VAD) framework is based on an encoder-decoder-style deep neural network, namely Spatio-Temporal Auto-Encoder (STAE).
The encoder and decoders in our model are designed with light-weight and simple structure, to meet the speed and computation limitations in real world VAD application.
The detailed STAE architecture and network parameters are depicted in Figure~\ref{fig:STAE}.
The target of STAE module is to learn the normal evolution patterns by encoding the spatial and temporal information among input frames in a normal video clip, so as to predict the future optical flows and the upcoming frames.
The whole model consists of an encoder and two separate decoders for predicting optical flows and RGB frames, respectively, as shown in Figure~\ref{fig:STAE}.

The designed encoder $\mathcal{E}$ for extracting spatio-temporal representations is constructed with a sequence of blocks including triple layers: 2D Convolution (Conv), batch-normalization (BN) and ReLU activation.
The kernel size of Conv layers is set to $3\times3$. We apply strided convolution (Strided-Conv) for downsampling, instead of using pooling layer. Such parametric operation is expected to support the network finding an informative way to downsample the spatial resolution of feature maps.
To enrich the spatio-temporal representations derived from the auto-encoder, we take advantages of RGB frames and optical flows as appearance and motion cues.
Specifically, we design two separate decoders, one for predicting the future optical flow maps and the other for the upcoming RGB frames.
The architecture decoder is also a sequence of layer blocks that increases the spatial resolution while reducing the number of feature maps after each deconvolution (Transpose-Conv) layer.
Note that skip connections are widely adopted design in DAE~\cite{ronneberger2015u,liu2018future,nguyen2019anomaly}.
Here, we also follow this devise and add them in the appearance prediction branch.

In the branch of appearance decoder $\mathcal{D}_a$, given the input video clip with length $k$ of moment $t$: $I_{t-k-1},...,I_{t-1},I_{t}$, the $L_2$ distance is adopted as the intensity metric that forces the model to produce an image $\hat{I}_{t+1}$ with similar intensity for each pixel in the future ground-truth frame $I_{t+1}$.
The intensity term is then formulated as:
\begin{equation}
\label{int}
\mathcal{L}_\mathrm{int}(I,\hat{I}) =  \|{I}-\hat{I}\|^2_2.
\end{equation}
Along with the intensity term, we also add a gradient constraint to preserve the original
gradients (i.e., the sharpness) in the predicted image. The gradient loss is defined as:
\begin{align}
\label{grd}
\mathcal{L}_\mathrm{grd}(I,\hat{I})= & \sum_{i, j}\big|\big|| I_{i, j}-I_{i-1, j}|-| \hat{I}_{i, j}-\hat{I}_{i-1, j}|\big|\big|_1  \nonumber  \\
& +\big|\big|| I_{i, j}-I_{i, j-1}|-| \hat{I}_{i, j}-\hat{I}_{i, j-1}|\big|\big|_1,
\end{align}
where $i, j$ indicate spatial pixel ordinates in an image. Combined with the intensity term, the final loss function for the appearance branch derives as:
\begin{equation}
\label{eqn_a}
\mathcal{L}_\mathrm{app}=\mathcal{L}_\mathrm{int}(I,\hat{I})+\lambda_\mathrm{grd}\mathcal{L}_\mathrm{grd}(I,\hat{I}).
\end{equation}

Besides the appearance evolution, the motion pattern also plays a significant role for anomaly distinction. Here, we consider the unsupervised optical flow prediction manner~\cite{jason2016back,meister2018unflow}, which could suit different scenes greatly through the data-driven end-to-end learning and requires no manual annotations in advance.
In motion decoder $\mathcal{D}_m$, for the pixel located at $\bm{x}=(i_x,i_y)$, it utilizes an image warp function $\delta(\cdot,\cdot)$ to map the current frame $I_{t}$ to future frame $\tilde{I}_{t+1}$ based on the predicted flow map $\tilde{F}$ as:
\begin{equation}
\begin{aligned}
\tilde{I}_{t+1}(\bm{x}) = \delta(I_t, \tilde{F}(\bm{x})) = I_t(\bm{x} - \tilde{F}(\bm{x})),
\end{aligned}
\end{equation}
where $\bm{x}$ stands for the two-dimensional coordinates of pixel $x$: $(i_x, j_x)^T$. Since it usually exists sub-grids in the warped frame, the bilinear interpolation is utilized, specifically:
\begin{equation}
\tilde{I}_{t+1}(\bm{x}) = \sum_{{y} \in \bm{N}({z})} I_t(\bm{z}) (1 - |i_y-i_{z}|_1)(1 - |j_y-j_{z}|_1),
\end{equation}
where $\bm{z}=\bm{x}-\tilde{F}(\bm{x})=(i_{z}, j_{z})^T$ represents for the source pixel point to perform bilinear interpolation, and $N(z)$ is a set of four nearest pixel neighbors of $z$.

By pushing the warped frame $\tilde{I}_{t+1}$ to be close to the ground-truth future frame ${I}_{t+1}$, the model automatically learns the flow maps. In this way, we are freed from the heavy step of extracting optical flow maps before either training or testing phase.
Correspondingly, the loss term of the motion branch is similar with appearance branch as:
\begin{equation}
\label{eqn_m}
\mathcal{L}_\mathrm{mot}=  \mathcal{L}_\mathrm{int}(I,\tilde{I})+\lambda_\mathrm{grd}\mathcal{L}_\mathrm{grd}(I,\tilde{I}).
\end{equation}
Finally, the appearance and motion loss functions are combined as in the following formula:
\begin{equation}
\mathcal{L}_\mathrm{ae}=  \mathcal{L}_\mathrm{app}+\lambda_\mathrm{mot}\mathcal{L}_\mathrm{mot}.
\end{equation}

\subsection{Relation Learning}
\label{sec_rlm}

Previous methods mainly address the VAD problem in an evolution pattern modeling manner.
During inference, the ones differing from the patterns learned in training phase are regarded as anomalies.
However, it is incomplete because the object anomalous status of certain evolution pattern is highly correlated with the scene, or even the region.
In our approach, we introduce knowledge graph embedding method to concretize the normal relationships by analyzing and summarizing the consistence between the objects and the scene they act in.
After extracting the spatio-temporal feature maps from the STAE, we utilize them as embeddings to construct a fact tuple.
In detail, we regard the scene embedding and objects embedding as the head and tail entities, respectively, for that a scene may contain various groups of objects, so as their evolution patterns.
Inspired by the KGE approaches~\cite{bordes2014semantic,dettmers2018convolutional,liu2016probabilistic} that conduct relation matching in a convolutional network model, we further extend the typical deep KGE approach to model spatial pattern relations according to different regions in a scene.
\begin{algorithm}[!h]
	\caption{Fast extraction of moving object regions}
	\SetAlgoLined
	\KwIn{A $k$ length video clip $\{I_{1},...,I_{k-1},I_{k}\}$. }
	\KwOut{A mask set $\mathcal{M}$ of moving objects regions.}
	
	\emph{initialize two $256\times256$ map of $A$ and $B$ filled with 0 and 1, separately}\;
	\For {$i \in [1,2,3,...,k-2]$}{
		$A\leftarrow A + abs((I_{i+2}-I_{i+1})-(I_{i+1}-I_{i}))$\;\tcp*[h]{$abs():$ get absolute value}  }
	apply Gaussian Blur on $A$ with kernel size of $5\times5$\;
	extract edges $E$ from $A$ with Sobel operator of kernel size $5$\;
	\For {$i \in [1,2,3,...,k-1]$}{
		$B\leftarrow B + abs(I_{i+1}-I_{i})$\; }
	$B\leftarrow B > 0.1$ \; \tcp*[h]{segment regions with threshold}\\
	do corrosion and expansion on $E*B$ with kernel size of $8\times8$ filled with $1$\;
	extract connected region set $\mathcal{G}$ in masked map $E*B$ with connectivity $=8$ \;
	\For {region map $ G \in\mathcal{G}$}{
		\If {$ G$\_width $>8$ and $ G$\_height $>8$}{
			generate a mask map $M$ with a rectangle covering ${G}$ with one and a half times size\;
			append $M$ to $\mathcal{M}$\;
		}
	}
	\label{algo:SegRule}
\end{algorithm}

A critical step lies in obtaining the embeddings for the objects and scenes as it might cause great computation burdens for object extraction. In this work, to facilitate the practical applications, we hereby propose a simple yet very effective mechanism for efficient extraction of moving objects with nearly ignorable computation consumption.
Under several traditional image processing steps like corrosion, expansion and connected regions extraction, we can easily obtain masks conform to regions of moving objects. 
The segmentation rule is based on processing pixel-distance maps among consecutive frames that clearly shows the moving area of objects. 
Further more, we propose to leverage 2nd-order pixel-distance map (acceleration) to eliminate noises coming from moving cameras, for that the pixel motion caused by moving cameras are typically smooth and uniform. The overall process is shown in Algorithm~\ref{algo:SegRule}.

After pre-processing, rectangle region masks covering the moving objects are prepared along with input video frames for extracting object and scene representations.
It's worthy mentioning that the anomalies are mostly related to moving objects with relatively rare static appearance outliers which can be easily figured out in general frame prediction.
Our segmentation rule is also beneficial for the following normal relation learning process as it can generate both object regions of the single and group pattern.
Thereafter, we obtain the frames of certain moving object(s) by segmenting out its region with the generated mask.
Instead of cropping the regions, we simply mask out the regions for preserving the relative object scale information, since the area of moving object regions vary largely with the number of objects therein.

Formally, given the output representations from STAE encoder with the resolution $w, h$ of $1/8$ raw inputs, we take the entire feature map as the scene embedding ${\rm E}_{s}$ and leverage the segment masks to extract moving object(s) embedding by computing ${\rm E}_{o}^{M}= {\rm E}_{s}*{\rm M}, \rm M \in \mathcal{M}$, where $\mathcal{M}$ is the set of segment masks, and ${\rm E}_{o}, {\rm E}_{s} \in \mathbb{R}^{d\times w\times h}$.
We preserve the spatial resolution of ${\rm E}_{s}$ and initialize a relation embedding of $d$ dimensions for each pixel location. In other word, the relation embedding map has the same size with the scene embedding. We denote the relation embedding map with ${\rm R} \in \mathbb{R}^{d\times w\times h}$. In the feed-forward pass, the relation learning module first concatenates input embeddings of ${\rm E}_{s}$ and the randomly initialized relation embedding map ${\rm R}$ ready to be learned, and feeds them through two 2D conv layers with filter ${\rm w}_1$ and ${\rm w}_2$, where the kernel size is set to 1 and each layer is followed by a ReLU unit and a batch-normalization unit.
After fusing relations, a feature map tensor $\Gamma \in \mathbb{R}^{d\times w\times h}$ is generated.
Finally, the obtained product $\Gamma$ is matched with the object embedding ${\rm e}_o^M \in \mathbb{R}^{d}$ via an inner product, where $e^{M}_o$ comes from the consensus of ${\rm E}^{M}_{o}$ with a global average pooling operation.
The process is depicted in Figure~\ref{fig:arch}.
Formally, the scoring function that measuring the plausibility of the normal relation between the object(s) and corresponding scene is defined as:
\begin{equation}
\psi_r({\rm E}_s,{\rm e}^{M}_o)=\sigma(\text{conv}_{{\rm w}_2}(\text{conv}_{{\rm w}_1}([{\rm E}_s;{\rm R}])) \cdot {\rm e}^{M}_o),
\end{equation}
where the dot product $\cdot$ uses the expanding operation for $e_o^M$ for spatial matching, and the final score map is normalized with a sigmoid function $\sigma$. Note that the relation embedding map $\rm R$ needs to be learnt during training.

For training the parameters, we propose a negative sample generation strategy and cooperate it with the mutual information maximization theory~\cite{hjelm2019learning}.
Two kinds of pseudo anomaly data are created by randomly speeding and ordering. In random speed strategy, we retain the first sampled frame and gradually add $k-1$ randomly chosen succeeding frames.
The intermittent frames are chosen by skipping frames using random intervals in the range of $\{1, 2, 3, 4\}$.
In random order strategy, we create the sample by switching the order of original frames.
By passing the above two groups of frames through the STAE encoder and the global average pooling layer, we can obtain their embeddings as $\acute{\rm e}^M_o, \grave{\rm e}^M_o$ w.r.t speeding sample and disordering sample, respectively.
The loss objective of relation learning is to maximize the mutual information of normal (positive) samples among all positive and negative samples:
\begin{equation}
\mathcal{L}_{\rm rl}= -\frac{1}{whn} \log \sum \limits_{\rm M\in\mathcal{M}} \sum \limits_{i, j}^{w,h}({\rm M}_{i,j}*\frac{\psi_r^{i,j}({\rm E}_s,{\rm e}^{M}_o)}{\sum \limits_{e \in {\{{\rm e}^{M}_o,\acute{\rm e}_o^{M},\grave{\rm e}^{M}_o\}}}\psi_r^{i,j}({\rm E}_s, e)}),
\end{equation}
where the region mask map $\rm M \in \mathbb{R}^{w\times h}$ is used to exclude object(s)-irrelevant areas and $n=|\mathcal{M}|$ denotes the number of region mask maps in a video clip.
Finally, the overall objective of the STRL framework is derived as:
\begin{equation}
\mathcal{L}=  \mathcal{L}_{\rm ae}+\lambda_{\rm rl}\mathcal{L}_{\rm rl}.
\end{equation}

\subsection{Anomaly Detection}
\label{sec_ad}

Given an input sequence of frames $I_{t-k},...,I_{t-1},I_{t}$ during the testing phase, we use the trained model to predict the next frame $\hat{I}_{t+1}$ and warped frame $\tilde{I}_{t+1}$ with the optical flow $\tilde{F}$.
Both the prediction results are compared with the ground-truth future frame $I_{t+1}$ by calculating the residual error ${S}_{\rm app}$ and ${S}_{\rm mot}$ according to Eqn.~(\ref{eqn_a}) and Eqn.~(\ref{eqn_m}). At the same time, anomaly relation scores are generated from the relation learning module, and we employ the minimum value among the relation score maps of a frame as the anomalous relation measurement, formally,
\begin{equation}
{S}_{\rm rl}= \min_{M \in\mathcal{M}}(\sum \limits_{i, j} {\rm M}_{i,j}*\psi_r^{i,j}({\rm E}_s,{\rm e}^{ M}_o)).
\end{equation}
After normalizing ${S}_{r}$, ${S}_{a}$ and ${S}_{m}$ of each video into range of [0,1], we obtain the final anomaly score for each clip:
\begin{equation}
{S} = {S}_{\rm app}+ \lambda_{\rm mot}{S}_{\rm mot} + \lambda_{\rm rl} {S}_{\rm rl}.
\end{equation}
We use the overall anomaly score $S$ to determine whether a frame/clip is anomalous or not.

\section{Experiments}
\label{sec:exp}
\begin{table}[t]
	\captionsetup{font={small}}
	\small
	\begin{center}
		\caption{Quantitative comparison with state-of-the-art methods for video anomaly detection. We measure the average AUC~(\%) on UCSD Ped2~\cite{li2013anomaly}, Avenue~\cite{lu2013abnormal}, and ShanghaiTech~\cite{luo2017revisit}. Numbers in bold indicate the best performance and the underscored ones are the second best. \textit{* denotes the evaluation under adjusted testing set on Avenue. The corresponding explanation is put in Sec~\ref{VsSOTA}, please refer to the statements with slanted font.}}
		\scalebox{0.8}{
			\begin{tabular}{ c | c | c | c | c | c}
				\hline
				Year & Methods & Ped1 & Ped2 & Avenue & Shanghai \\
				\hline
				\multirow{5}{*}{\rotatebox{90}{before 2016\hspace{0.1cm}}}
				&Mehran \etal~\cite{mehran2009abnormal} & - & 55.6 & - & - \\
				&Kim \etal~\cite{kim2009observe} & 59.0 & 69.3 & - & - \\
				&Mahadevan \etal~\cite{mahadevan2010anomaly} & 81.8 & 82.9 & - & - \\
				&Lu \etal~\cite{lu2013abnormal} & - & - & 80.9 & - \\
				&Xu \etal~\cite{xu2015learning} & 67.2 & 90.8 & - & \\
				\hline
				\multirow{5}{*}{\rotatebox{90}{2017-2016\hspace{0.1cm}}}
				&Hasan \etal~\cite{hasan2016learning} & 75.0 & 85.0 & 80.0 & 60.9 \\
				&Hinami \etal~\cite{hinami2017joint} & - & 92.2 & - & - \\
				&Tudor \etal~\cite{tudor2017unmasking} & 68.4 & 82.2 & 80.6 & - \\
				&Ravanbakhsh \etal~\cite{ravanbakhsh2017abnormal} & 70.3 & 93.5 & - & - \\
				&Luo \etal~\cite{luo2017revisit} & - & 92.2 & 81.7 & 68.0 \\
				&Xu \etal~\cite{xu2017detecting} & - & 90.8 & - & - \\
				\hline
				\multirow{10}{*}{\rotatebox{90}{2021-2018\hspace{0.1cm}}}
				&Liu \etal~\cite{liu2018future}  & 83.1 & 95.4 & 85.1 & 72.8 \\
				&Nguyen \etal~\cite{nguyen2019anomaly} & - & 96.2 & 86.9 & -\\
				&Gong \etal~\cite{gong2019memorizing} & - & 94.1 & 83.3 & 71.2 \\	
				&Pang \etal~\cite{pang2020self} & 71.7 & 83.2 & - & -\\
				&Dong \etal~\cite{dong2020dual} & - & 95.6 & 84.9 & 73.7  \\
				&Tang \etal~\cite{tang2020integrating} & 82.6 & 96.3 & 85.1 & 73.0 \\	
				&Zaheer \etal~\cite{zaheer2020old} & - & \underline{98.1} & - & - \\	
				&Park \etal~\cite{park2020learning} & - & {97.0}  & \underline{88.5}(87.9*)&  70.5\\
				&Lv \etal~\cite{lv2021learning} & 85.1& {96.9}  & \textbf{89.5} (88.7*)&  \underline{73.8}\\
				\cline{2-5}
				&STRL. & \textbf{87.1} & \textbf{99.0}  &  88.7 ({91.3*}) &  \textbf{76.0}\\
				\hline
		\end{tabular}}
		\label{table:Comparison}
	\end{center}
	
\end{table}
\subsection{Datasets.}
\label{Dataset}
To evaluate both qualitative and quantitative results of the proposed method and compare it with state-of-the-art algorithms, we conduct experiments on three public video anomaly detection benchmarks, {\em i.e.}, UCSD Ped2 \cite{mahadevan2010anomaly}, CUHK Avenue \cite{lu2013abnormal} and ShanghaiTech~\cite{luo2017revisit}.

\noindent \textbf{Ped 1\&2.} The UCSD Ped1\& Ped2 dataset \cite{li2013anomaly} contains 34 and 16 training videos, 36 and 12 test videos, respectively, with 12 irregular events. Examples of abnormal events are bikers, skaters and cars in a pedestrian area. The resolution of each video is $240\times360$ pixels.

\noindent \textbf{Avenue.} The CUHK Avenue dataset \cite{lu2013abnormal} consists of 16 training and 21 test videos. Examples of anomalous events in Avenue are related to people running, throwing objects or walking in the wrong direction. The resolution of each video is $360\times640$ pixels.

\noindent \textbf{ShanghaiTech.} The ShanghaiTech dataset \cite{luo2017revisit} is a very challenging benchmark that contains videos from 13 scenes with complex light conditions and camera angles. The overall number of frames for training and testing reach up to 274K and 42K, respectively. The training videos contain only normal events, while the test videos contain normal and abnormal sequences. Examples of anomalous events are: robbing, jumping, fighting and bikers in pedestrian areas.
The resolution of each video is $480\times856$ pixels.

\subsection{Implementation details}
\label{Setup}
\noindent \textbf{Evaluation Metrics.}
Following prior works~\cite{liu2018future,luo2017remembering,mahadevan2010anomaly}, we evaluate the performance using the area under ROC curve (AUC). 
ROC curve is obtained by varying the threshold for the anomaly score for each frame-wise prediction. Higher AUC score indicates better VAD accuracy.

\noindent \textbf{Implementation Details.} Our model takes as input video frames with the resolution of $256 \times 256$ and normalizes them to the range of $[-1,1]$. 
All training epochs are set to $1000$ on Ped2, Avenue and ShanghaiTech, with learning rate $0.00001$ and batch size $4$. 
The balance weights in the objective functions and anomaly score production are set as $\lambda_{grd} =1$, $\lambda_{mot} =1$ and $\lambda_{rl} =0.5$. The final anomaly scores are processed with mean filter of size $15$ for smoothness.

\begin{figure*}
	\centering
	\includegraphics[width=1\textwidth]{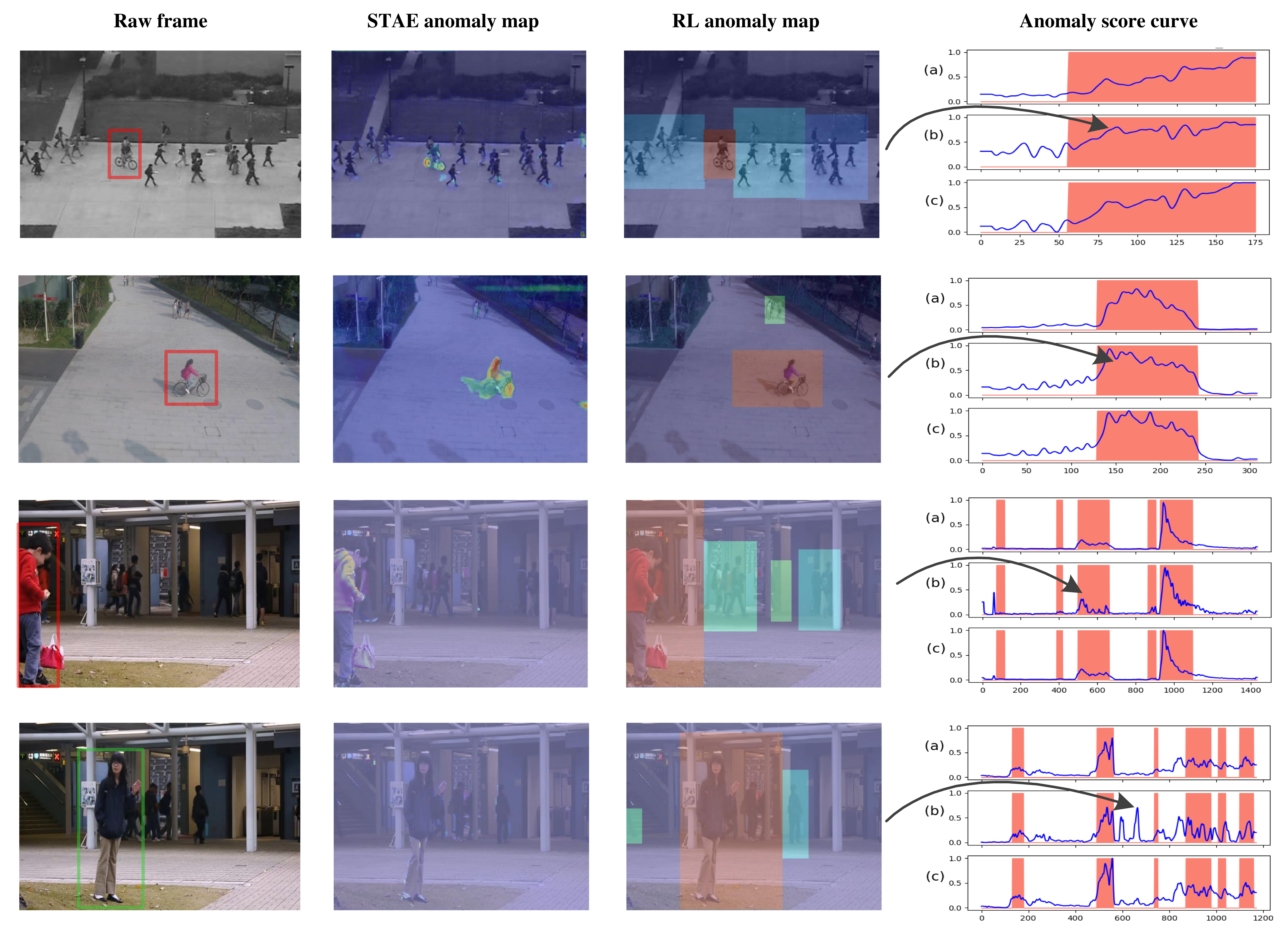}
	\caption{Visualization of some test examples on various benchmarks. The listed four rows are the selected samples from Ped2, ShanghaiTech and Avenue, respectively. The red boxes and green box in raw images cover the ground-truth anomalous and normal areas, respectively. In the columns of `STAE anomaly map' and `RL anomaly map', the highlighted area stand for the predicted anomaly region. Color depth is related to the degree of abnormality. The darker the color, the higher the degree of abnormality. In each subfigure of the anomaly score curves, the blue curve is the predicted anomaly scores of consecutive frames and the red area represents the temporal ground-truth anomaly locations. The (a), (b) and (c) rows belong to the results of STAE module, RL module and the whole STRL framework, respectively, with the x-axis denotes the frame number in a video sequence and the y-axis denotes the scalar of anomaly scores. Best viewed in color.} 
	\label{fig:Output}
\end{figure*}

\subsection{Comparisons with SOTA Methods} 
\label{VsSOTA}
To show the effectiveness of our method, we compare the proposed STRL framework with many well-designed reconstruction based state-of-the-art methods since 2018, as well as early methods based on hand-crafted or deep features (before 2016 and 2016$\sim$2018, respectively).
In Table~\ref{table:Comparison} we present the comparative results of our STAE module and the entire frameworks versus the previous literature.
The sanity check is conducted under the standard setup with the training set and testing set provided as the original datasets.
Thus all methods can be directly put in comparison.
We report the frame-level AUC scores (whenever available) on three widely adopted benchmarks: ShanghaiTech, Avenue and UCSD Ped1\&2.
The superior performance demonstrates the effectiveness of our STRL framework.
Specifically, our model surpasses previous state-of-the-art model (Lv \etal~\cite{lv2021learning}) by a large margin, which is up to $2\%$ on ShanghaiTech and over $2\%$ on Ped2.
Visible test samples are shown in Figure~\ref{fig:Output}. 
The results show that our relation learning technology not only works well on simple scenes like Ped 1\&2, but also performs robustly on complex scenes as ShanghaiTech.
It's worth mention that the proposed fast moving region extraction and introduction of image warp function free our approach from heavy pre-processing steps like optical flow map extraction, and make our approach more applicable in practice. Further efficiency analysis is conducted in Sec.\ref{MC&IS}.

It is showed that on Avenue, our model performs sightly worse than the previous best approaches. According to failure cases analysis, it is mainly because there exist some ambiguous objects in this benchmark. A case is shown in the fourth row of Figure~\ref{fig:Output}. The woman standing on the lawn with small-scale movements is regarded as normal, however person walking or playing in the same area is anomaly as in the 3rd row. The benchmark annotators simply set the close-to-static objects as the normal and ignored these minor abnormalities. In this manner, the future prediction approaches take advantages over other methods that they usually can well predict the objects with small or no motion variation. However, this setting is unreliable in real applications.
On the contrary, our model is able to capture the objects with small-scale movements and distinguish whether they are expected to appear in the scene (region). 

In order to better measure the effectiveness of the RL module, we simply remove the videos contain ambiguous objects in the testing set, and then evaluate our model, as well as the previous best model (Lv \etal~\cite{lv2021learning} and Park \etal~\cite{park2020learning}) under the new setting. The adjusted data split will be made public together with the implementation codes of our framework.
In conclusion, the state-of-the-art performance on the adjusted Avenue testing set shows that our model is more comprehensive and sensitive for distinguishing anomalies of unexpected object-scene relation.

\subsection{Ablation Study} 
\label{AbS}
In this section, we conduct ablation studies for analyzing the effectiveness of each component in the STAE module and evaluating the role of relation learning for VAD.
The comparison between the predicted anomaly score curves of models with different setting are shown in Figure~\ref{fig:Output}.
\subsubsection{Analysis of the STAE module} 
\label{AS_STAE}
We first analyze the effectiveness of the STAE components.
The quantitative results are listed in Table~\ref{table:STAE}. As it could be seen, solely utilizing the appearance or motion cues could lead to satisfactory anomaly detection results since in most cases, merely the appearance or motion information could already provide reliable judgment. The non-trivial anomaly detection results of the motion branch also indicate the effectiveness of the unsupervised learning for the optical flows. Further, combining both the appearance and motion branch could lead to better results. It is because in some challenging cases, both the appearance and motion cues are important for effective anomaly identification.
In essence, the aggregation of predicting the future frames and optical flows urge the AE model to capture the complementary aspects of both the spatial and temporal evolution patterns for normal events.
Therefore, a more comprehensive representation is learned and encoded in the model, which is beneficial for latter relation analysis.
\begin{figure}[t]
	\begin{center}
		\begin{picture}(240,170)
		\put(0,90){\includegraphics[scale=0.33]{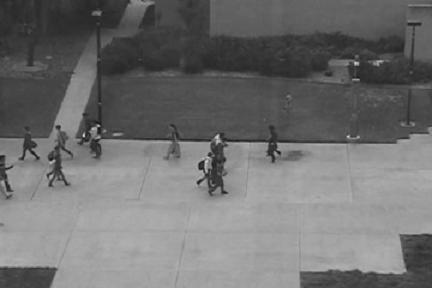}}
		\put(120,90){\includegraphics[scale=0.25]{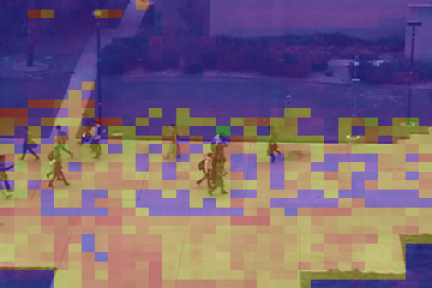}}
		\put(0,0){\includegraphics[scale=0.25]{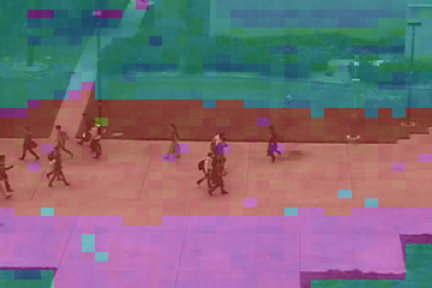}}
		\put(120,0){\includegraphics[scale=0.25]{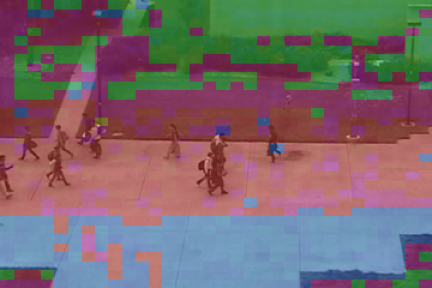}}
		\put(36,80){raw image}\put(163,80){$c=2$}\put(43,-10){$c=3$}\put(163,-10){$c=4$}
		\end{picture}
	\end{center}
	\caption{Similarity analysis of concretized relation embeddings. K-means is leveraged to automatically cluster the embeddings into $c$ class. Various colors are assigned to different classes with the degree corresponding with the similarity to the class centroids.}
	\label{fig:Simi}
\end{figure}
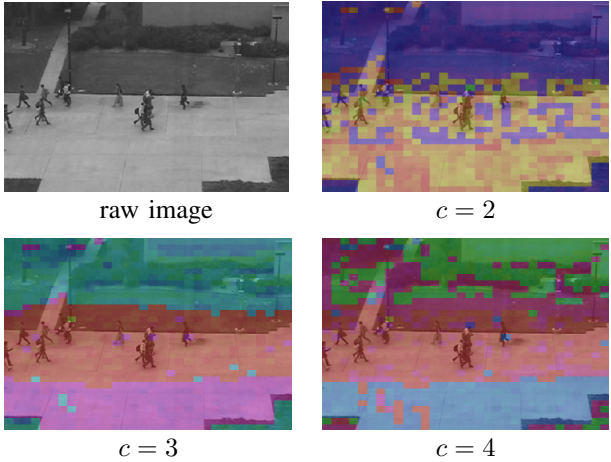

\begin{table}[t]
	\captionsetup{font={small}}
	\small
	\begin{center}
		\caption{Effectiveness analysis of the RL module. `Avenue*' denotes the AUC results on the adjusted testing set.}
		\scalebox{0.85}{
			\begin{tabular}{c | c  c  c  c  c}
				\hline
				Setting & ShanghaiTech & Avenue & Avenue* & Ped2 & Ped1\\
				\hline
				STAE. & 73.8 & 86.0 & 85.9 & 96.5 & 85.3\\
				RL. & 74.9 & 84.7 & 88.9 & 98.0 & 83.1\\
				\hline
				STRL. & 76.0 & 88.7 & 91.3 & 99.0 & 87.1\\
				\hline
		\end{tabular}}
		\label{table:RL}
	\end{center}
\end{table}
\subsubsection{Analysis of the RL module} 
\label{AS_RL}
In this part, we evaluate the effectiveness of the Relation Learning module.
As shown in Table~\ref{table:RL}, the introduction of KGE brings about over $2\%$ AUC gain across all benchmarks.
It is because our RL module is devised to look into anomaly analyzing from a brand new perspective.
We take into consideration the relations between the pattern of moving object(s) and the corresponding scene area, specific to the moving area of the object.
Then tackling the VAD problem in a high-level semantic space.

In addition, it is interesting  to understand how the relation embedding works and what the information it contains. 
To this end, we analyze the similarity of the relation embeddings in spatial domain.
Concretely, we perform the K-means technology on the relation embeddings concretized from Ped2 training set.
Since Ped2 is a single scene benchmark, the relation embedding on a certain spatial location only focuses on the relation patterns of a certain region.
The clustering results are visualized in Figure~\ref{fig:Simi}, where the corresponding clusters with the cluster number ranging from $2$ to $4$ are showed, together with a raw image from Ped2.
It is obvious that the relation embeddings are automatically aggregated into several meaningful classes. 
As in Subfigure $(c=3)$, the three clusters correspond to the moving areas of people, the region where few people pass by, and the distant view of lawn and buildings.
A similar tendency can be seen in the Subfigure $(c=2)$.
When $c$ gets larger, the lawn and the buildings are additionally separated.
The visualization demonstrates that the designed RL module is able to collect region-wise normal relation patterns.

\vspace{-0.3cm}
\subsection{Model Complexity and Inference Speed}
\label{MC&IS}
As in Table~\ref{table:Params&Speed}, our model is extremely light-weight when it comes to the model parameters, almost $1/70$ the amount of Lv \etal~\cite{lv2021learning}. In addition, our model runs at a fast speed of $180+$ FPS on a Nvidia RTX-2080Ti GPU. 
In summary, our end-to-end model achieves a good performance with a small consumption of memory and computation. The fast inference speed and light weight design makes our method more favorable in real-world applications.

\begin{table}[t]
	\captionsetup{font={small}}
	\small
	\begin{center}
		\caption{Effectiveness analysis of the STAE components. `Appearance' and `Motion' denote the branches of the future frames and optical flows predicting, respectively.}
		
		\scalebox{0.85}{
			\begin{tabular}{c | c  c  c  c  c}
				\hline
				Setting & ShanghaiTech & Avenue & Avenue* & Ped2 & Ped1\\
				\hline
				Appearance & 72.6 & 85.5 & 84.5 & 94.1 & 83.3\\
				Motion & 73.1 & 83.9 & 82.3 & 95.5 & 84.7\\
				\hline
				STAE. & 73.8 & 86.0 & 85.9 & 96.5 & 85.3\\
				\hline
		\end{tabular}}
		\label{table:STAE}
	\end{center}
\end{table}
\begin{table}[h]
	\captionsetup{font={small}}
	\small
	\begin{center}
		\caption{Analysis on the model complexity and inference speed of various SOTA methods. The inference speed information is collected by running the official implements on a single Nvidia RTX-2080Ti GPU.}
		\begin{tabular}{c | c | c}
			\hline
			Methods & Parameters (M) & FPS \\
			\hline
			Lu \etal~\cite{lu2020few}& 19.0 & 2.0 \\
			Gong \etal~\cite{gong2019memorizing} & 6.2 & 80.7 \\
			Park \etal~\cite{park2020learning}& 15.0 &  121.7\\
			Lv \etal~\cite{lv2021learning} & 12.7 &  160.1\\
			\hline
			Ours& 0.18 &  183.8\\
			\hline
		\end{tabular}
		\label{table:Params&Speed}
	\end{center}
\end{table}
\section{Conclusion}
In this work, the object-scene relations are specifically considered for unsupervised anomaly detection which is ignored by previous works. To this end, a Spatial-Temporal Relation Learning (STRL) framework is proposed for explicit modeling the normal relations within the videos. First, a Spatial-Temporal Auto-Encoder (STAE) is designed for capturing the spatial-temporal evolution patterns through the joint learning of an appearance and a motion branch. Based on the learned representations, a Relation Learning (RL) module is proposed to explicitly formalize the normal relation learning via an introduced knowledge graph embedding method. An optimizable object-scene relation score map is learned for effectively summarizing the relationship between the object and scene embeddings.  Extensive experimental evaluations demonstrate the efficiency of the STAE module and the significance of relation learning for VAD. In conclusion, Relation modeling is a potential direction for VAD, which deserves further investigation and research.

{
\bibliographystyle{IEEEtran}
\bibliography{egbib}

\begin{thebibliography}{10}
\providecommand{\url}[1]{#1}
\csname url@samestyle\endcsname
\providecommand{\newblock}{\relax}
\providecommand{\bibinfo}[2]{#2}
\providecommand{\BIBentrySTDinterwordspacing}{\spaceskip=0pt\relax}
\providecommand{\BIBentryALTinterwordstretchfactor}{4}
\providecommand{\BIBentryALTinterwordspacing}{\spaceskip=\fontdimen2\font plus
\BIBentryALTinterwordstretchfactor\fontdimen3\font minus
  \fontdimen4\font\relax}
\providecommand{\BIBforeignlanguage}[2]{{%
\expandafter\ifx\csname l@#1\endcsname\relax
\typeout{** WARNING: IEEEtran.bst: No hyphenation pattern has been}%
\typeout{** loaded for the language `#1'. Using the pattern for}%
\typeout{** the default language instead.}%
\else
\language=\csname l@#1\endcsname
\fi
#2}}
\providecommand{\BIBdecl}{\relax}
\BIBdecl

\bibitem{ramachandra2020survey}
B.~Ramachandra, M.~Jones, and R.~R. Vatsavai, ``A survey of single-scene video
  anomaly detection,'' \emph{IEEE Transactions on Pattern Analysis and Machine
  Intelligence}, 2020.

\bibitem{kiran2018overview}
B.~R. Kiran, D.~M. Thomas, and R.~Parakkal, ``An overview of deep learning
  based methods for unsupervised and semi-supervised anomaly detection in
  videos,'' \emph{Journal of Imaging}, 2018.

\bibitem{chandola2009anomaly}
V.~Chandola, A.~Banerjee, and V.~Kumar, ``Anomaly detection: A survey,''
  \emph{ACM computing surveys (CSUR)}, 2009.

\bibitem{santhosh2020anomaly}
K.~K. Santhosh, D.~P. Dogra, and P.~P. Roy, ``Anomaly detection in road traffic
  using visual surveillance: A survey,'' \emph{ACM Computing Surveys (CSUR)},
  2020.

\bibitem{hasan2016learning}
M.~Hasan, J.~Choi, J.~Neumann, A.~K. Roy-Chowdhury, and L.~S. Davis, ``Learning
  temporal regularity in video sequences,'' in \emph{CVPR}, 2016.

\bibitem{luo2017revisit}
W.~Luo, W.~Liu, and S.~Gao, ``A revisit of sparse coding based anomaly
  detection in stacked rnn framework,'' in \emph{ICCV}, 2017.

\bibitem{liu2018future}
W.~Liu, W.~Luo, D.~Lian, and S.~Gao, ``Future frame prediction for anomaly
  detection--a new baseline,'' in \emph{CVPR}, 2018.

\bibitem{nguyen2019anomaly}
T.-N. Nguyen and J.~Meunier, ``Anomaly detection in video sequence with
  appearance-motion correspondence,'' in \emph{ICCV}, 2019.

\bibitem{gong2019memorizing}
D.~Gong, L.~Liu, V.~Le, B.~Saha, M.~R. Mansour, S.~Venkatesh, and A.~v.~d.
  Hengel, ``Memorizing normality to detect anomaly: Memory-augmented deep
  autoencoder for unsupervised anomaly detection,'' in \emph{ICCV}, 2019.

\bibitem{park2020learning}
H.~Park, J.~Noh, and B.~Ham, ``Learning memory-guided normality for anomaly
  detection,'' in \emph{CVPR}, 2020.

\bibitem{lv2021localizing}
H.~Lv, C.~Zhou, Z.~Cui, C.~Xu, Y.~Li, and J.~Yang, ``Localizing anomalies from
  weakly-labeled videos,'' \emph{IEEE transactions on image processing}, 2021.

\bibitem{lv2021learning}
H.~Lv, C.~Chen, Z.~Cui, C.~Xu, Y.~Li, and J.~Yang, ``Learning normal dynamics
  in videos with meta prototype network,'' in \emph{CVPR}, 2021.

\bibitem{masci2011stacked}
J.~Masci, U.~Meier, D.~Cire{\c{s}}an, and J.~Schmidhuber, ``Stacked
  convolutional auto-encoders for hierarchical feature extraction,'' in
  \emph{ICANN}, 2011.

\bibitem{sabokrou2016video}
M.~Sabokrou, M.~Fathy, and M.~Hoseini, ``Video anomaly detection and
  localization based on the sparsity and reconstruction error of
  auto-encoder,'' \emph{Electronics Letters}, 2016.

\bibitem{chalapathy2017robust}
R.~Chalapathy, A.~K. Menon, and S.~Chawla, ``Robust, deep and inductive anomaly
  detection,'' in \emph{Joint European Conference on Machine Learning and
  Knowledge Discovery in Databases}, 2017.

\bibitem{sabokrou2018adversarially}
M.~Sabokrou, M.~Khalooei, M.~Fathy, and E.~Adeli, ``Adversarially learned
  one-class classifier for novelty detection,'' in \emph{CVPR}, 2018.

\bibitem{abati2019latent}
D.~Abati, A.~Porrello, S.~Calderara, and R.~Cucchiara, ``Latent space
  autoregression for novelty detection,'' in \emph{CVPR}, 2019.

\bibitem{lu2019future}
Y.~Lu, M.~Kumar Krishna~Reddy, S.~s. Nabavi, and Y.~Wang, ``Future frame
  prediction using convolutional vrnn for anomaly detection,'' in \emph{AVSS},
  2019.

\bibitem{luo2017remembering}
W.~Luo, W.~Liu, and S.~Gao, ``Remembering history with convolutional lstm for
  anomaly detection,'' in \emph{ICME}, 2017.

\bibitem{liu2021hybrid}
Z.~Liu, Y.~Nie, C.~Long, Q.~Zhang, and G.~Li, ``A hybrid video anomaly
  detection framework via memory-augmented flow reconstruction and flow-guided
  frame prediction,'' in \emph{ICCV}, 2021.

\bibitem{ilg2017flownet}
E.~Ilg, N.~Mayer, T.~Saikia, M.~Keuper, A.~Dosovitskiy, and T.~Brox, ``Flownet
  2.0: Evolution of optical flow estimation with deep networks,'' in
  \emph{CVPR}, 2017.

\bibitem{hur2021survey}
A.~Hur, N.~Janjua, and M.~Ahmed, ``A survey on state-of-the-art techniques for
  knowledge graphs construction and challenges ahead,'' \emph{arXiv preprint
  arXiv:2110.08012}, 2021.

\bibitem{dettmers2018convolutional}
T.~Dettmers, P.~Minervini, P.~Stenetorp, and S.~Riedel, ``Convolutional 2d
  knowledge graph embeddings,'' in \emph{Thirty-second AAAI conference on
  artificial intelligence}, 2018.

\bibitem{wang2017knowledge}
Q.~Wang, Z.~Mao, B.~Wang, and L.~Guo, ``Knowledge graph embedding: A survey of
  approaches and applications,'' \emph{IEEE Transactions on Knowledge and Data
  Engineering}, 2017.

\bibitem{adam2008robust}
A.~Adam, E.~Rivlin, I.~Shimshoni, and D.~Reinitz, ``Robust real-time unusual
  event detection using multiple fixed-location monitors,'' \emph{TPAMI}, 2008.

\bibitem{antic2011video}
B.~Anti{\'c} and B.~Ommer, ``Video parsing for abnormality detection,'' in
  \emph{ICCV}, 2011.

\bibitem{lu2013abnormal}
C.~Lu, J.~Shi, and J.~Jia, ``Abnormal event detection at 150 fps in matlab,''
  in \emph{ICCV}, 2013.

\bibitem{mahadevan2010anomaly}
V.~Mahadevan, W.~Li, V.~Bhalodia, and N.~Vasconcelos, ``Anomaly detection in
  crowded scenes,'' in \emph{CVPR}, 2010.

\bibitem{mehran2009abnormal}
R.~Mehran, A.~Oyama, and M.~Shah, ``Abnormal crowd behavior detection using
  social force model,'' in \emph{CVPR}, 2009.

\bibitem{piciarelli2008trajectory}
C.~Piciarelli, C.~Micheloni, and G.~L. Foresti, ``Trajectory-based anomalous
  event detection,'' \emph{IEEE Transactions on Circuits and Systems for video
  Technology}, 2008.

\bibitem{zhang2009learning}
T.~Zhang, H.~Lu, and S.~Z. Li, ``Learning semantic scene models by object
  classification and trajectory clustering,'' in \emph{CVPR}, 2009.

\bibitem{cong2011sparse}
Y.~Cong, J.~Yuan, and J.~Liu, ``Sparse reconstruction cost for abnormal event
  detection,'' in \emph{CVPR}, 2011.

\bibitem{cheng2015video}
K.-W. Cheng, Y.-T. Chen, and W.-H. Fang, ``Video anomaly detection and
  localization using hierarchical feature representation and gaussian process
  regression,'' in \emph{CVPR}, 2015.

\bibitem{zhao2011online}
B.~Zhao, L.~Fei-Fei, and E.~P. Xing, ``Online detection of unusual events in
  videos via dynamic sparse coding,'' in \emph{CVPR}, 2011.

\bibitem{kim2009observe}
J.~Kim and K.~Grauman, ``Observe locally, infer globally: a space-time mrf for
  detecting abnormal activities with incremental updates,'' in \emph{CVPR},
  2009.

\bibitem{5206569}
------, ``Observe locally, infer globally: A space-time mrf for detecting
  abnormal activities with incremental updates,'' in \emph{CVPR}, 2009.

\bibitem{lv2020localizing}
H.~Lv, C.~Zhou, C.~Xu, Z.~Cui, and J.~Yang, ``Localizing anomalies from
  weakly-labeled videos,'' \emph{ArXiv}, 2020.

\bibitem{berant2013semantic}
J.~Berant, A.~Chou, R.~Frostig, and P.~Liang, ``Semantic parsing on freebase
  from question-answer pairs,'' in \emph{EMNLP}, 2013.

\bibitem{heck2013leveraging}
L.~Heck, D.~Hakkani-T{\"u}r, and G.~Tur, ``Leveraging knowledge graphs for
  web-scale unsupervised semantic parsing,'' in \emph{ISCA}, 2013.

\bibitem{damljanovic2012named}
D.~Damljanovic and K.~Bontcheva, ``Named entity disambiguation using linked
  data,'' in \emph{ESWC}, 2012.

\bibitem{zheng2012entity}
Z.~Zheng, X.~Si, F.~Li, E.~Y. Chang, and X.~Zhu, ``Entity disambiguation with
  freebase,'' in \emph{WIIAT}, 2012.

\bibitem{hoffmann2011knowledge}
R.~Hoffmann, C.~Zhang, X.~Ling, L.~Zettlemoyer, and D.~S. Weld,
  ``Knowledge-based weak supervision for information extraction of overlapping
  relations,'' in \emph{AMACL}, 2011.

\bibitem{daiber2013improving}
J.~Daiber, M.~Jakob, C.~Hokamp, and P.~N. Mendes, ``Improving efficiency and
  accuracy in multilingual entity extraction,'' in \emph{ICSS}, 2013.

\bibitem{bordes2014open}
A.~Bordes, J.~Weston, and N.~Usunier, ``Open question answering with weakly
  supervised embedding models,'' in \emph{Joint European conference on machine
  learning and knowledge discovery in databases}, 2014.

\bibitem{bordes2014question}
A.~Bordes, S.~Chopra, and J.~Weston, ``Question answering with subgraph
  embeddings,'' \emph{EMNLP}, 2014.

\bibitem{ronneberger2015u}
O.~Ronneberger, P.~Fischer, and T.~Brox, ``U-net: Convolutional networks for
  biomedical image segmentation,'' in \emph{MICCAI}, 2015.

\bibitem{jason2016back}
J.~Y. Jason, A.~W. Harley, and K.~G. Derpanis, ``Back to basics: Unsupervised
  learning of optical flow via brightness constancy and motion smoothness,'' in
  \emph{ECCV}, 2016.

\bibitem{meister2018unflow}
S.~Meister, J.~Hur, and S.~Roth, ``Unflow: Unsupervised learning of optical
  flow with a bidirectional census loss,'' in \emph{AAAI}, 2018.

\bibitem{bordes2014semantic}
A.~Bordes, X.~Glorot, J.~Weston, and Y.~Bengio, ``A semantic matching energy
  function for learning with multi-relational data,'' \emph{Machine Learning},
  2014.

\bibitem{liu2016probabilistic}
Q.~Liu, H.~Jiang, A.~Evdokimov, Z.-H. Ling, X.~Zhu, S.~Wei, and Y.~Hu,
  ``Probabilistic reasoning via deep learning: Neural association models,''
  \emph{Arxiv}, 2016.

\bibitem{hjelm2019learning}
R.~D. Hjelm, A.~Fedorov, S.~Lavoie-Marchildon, K.~Grewal, P.~Bachman,
  A.~Trischler, and Y.~Bengio, ``Learning deep representations by mutual
  information estimation and maximization,'' \emph{ICLR}, 2019.

\bibitem{li2013anomaly}
W.~Li, V.~Mahadevan, and N.~Vasconcelos, ``Anomaly detection and localization
  in crowded scenes,'' \emph{TPAMI}, 2013.

\bibitem{xu2015learning}
D.~Xu, E.~Ricci, Y.~Yan, J.~Song, and N.~Sebe, ``Learning deep representations
  of appearance and motion for anomalous event detection,'' \emph{BMVC}, 2015.

\bibitem{hinami2017joint}
R.~Hinami, T.~Mei, and S.~Satoh, ``Joint detection and recounting of abnormal
  events by learning deep generic knowledge,'' in \emph{CVPR}, 2017.

\bibitem{tudor2017unmasking}
R.~Tudor~Ionescu, S.~Smeureanu, B.~Alexe, and M.~Popescu, ``Unmasking the
  abnormal events in video,'' in \emph{ICCV}, 2017.

\bibitem{ravanbakhsh2017abnormal}
M.~Ravanbakhsh, M.~Nabi, E.~Sangineto, L.~Marcenaro, C.~Regazzoni, and N.~Sebe,
  ``Abnormal event detection in videos using generative adversarial nets,'' in
  \emph{ICIP}, 2017.

\bibitem{xu2017detecting}
D.~Xu, Y.~Yan, E.~Ricci, and N.~Sebe, ``Detecting anomalous events in videos by
  learning deep representations of appearance and motion,'' \emph{CVIU}, 2017.

\bibitem{pang2020self}
G.~Pang, C.~Yan, C.~Shen, A.~v.~d. Hengel, and X.~Bai, ``Self-trained deep
  ordinal regression for end-to-end video anomaly detection,'' in \emph{CVPR},
  2020.

\bibitem{dong2020dual}
F.~Dong, Y.~Zhang, and X.~Nie, ``Dual discriminator generative adversarial
  network for video anomaly detection,'' \emph{IEEE Access}, 2020.

\bibitem{tang2020integrating}
Y.~Tang, L.~Zhao, S.~Zhang, C.~Gong, G.~Li, and J.~Yang, ``Integrating
  prediction and reconstruction for anomaly detection,'' \emph{Pattern
  Recognition Letters}, 2020.

\bibitem{zaheer2020old}
M.~Z. Zaheer, J.-h. Lee, M.~Astrid, and S.-I. Lee, ``Old is gold: Redefining
  the adversarially learned one-class classifier training paradigm,'' in
  \emph{CVPR}, 2020.

\bibitem{lu2020few}
Y.~Lu, F.~Yu, M.~K.~K. Reddy, and Y.~Wang, ``Few-shot scene-adaptive anomaly
  detection,'' in \emph{ECCV}, 2020.

\end{thebibliography}

\vfill

\end{document}